%% file: main.tex
\documentclass[10pt,twocolumn,letterpaper]{article}

\usepackage[pagenumbers]{cvpr}

\input{preamble}

\usepackage{graphicx}
\usepackage[pagebackref,breaklinks,colorlinks,citecolor=cvprblue]{hyperref}
\usepackage{xcolor}
\usepackage{amsmath}
\usepackage{amssymb}
\usepackage{amsfonts}
\usepackage{dsfont}
\usepackage{array}
\usepackage{multirow}
\usepackage{stfloats}
\usepackage{booktabs}
\usepackage[dvipsnames]{xcolor}

\input{preamble}
\definecolor{cvprblue}{rgb}{0.21,0.49,0.74}
\usepackage[pagebackref,breaklinks,colorlinks,citecolor=cvprblue]{hyperref}
 
\title{MIST: \underline{M}itigating \underline{I}nter\underline{s}ectional Bias with Disentangled Cross-Attention Editing in \underline{T}ext-to-Image Diffusion Models}

\author{Hidir Yesiltepe \qquad
Kiymet Akdemir  \qquad
Pinar Yanardag\\
Virginia Tech\\
{\tt\small \{hidir, kiymet, pinary\}@vt.edu} \\
\small{Project webpage: \url{https://mist-diffusion.github.io}}
}

\begin{document}

\input{figs/teaser_sec}
 
\maketitle

\input{sec/0_abstract} 
\input{sec/1_intro} 
\input{sec/2_related}

\input{sec/3_background}

\input{sec/4_methodology}

\input{sec/5_experiments}

\input{sec/6_conclusion}
{\small
\bibliographystyle{ieee_fullname}
\bibliography{egbib}
}

\input{sec/7_appendix}

\end{document}

%% file: preamble.tex
\usepackage[dvipsnames]{xcolor}

%% file: figs/teaser_sec.tex
\twocolumn[{
\maketitle
\begin{center}
    \captionsetup{type=figure}
    \vspace{-1em}
\newcommand{\imwidth}{0.88\textwidth}

\begin{tabular}{@{}c@{}}
 
\parbox{\imwidth}{\includegraphics[width=\imwidth, ]{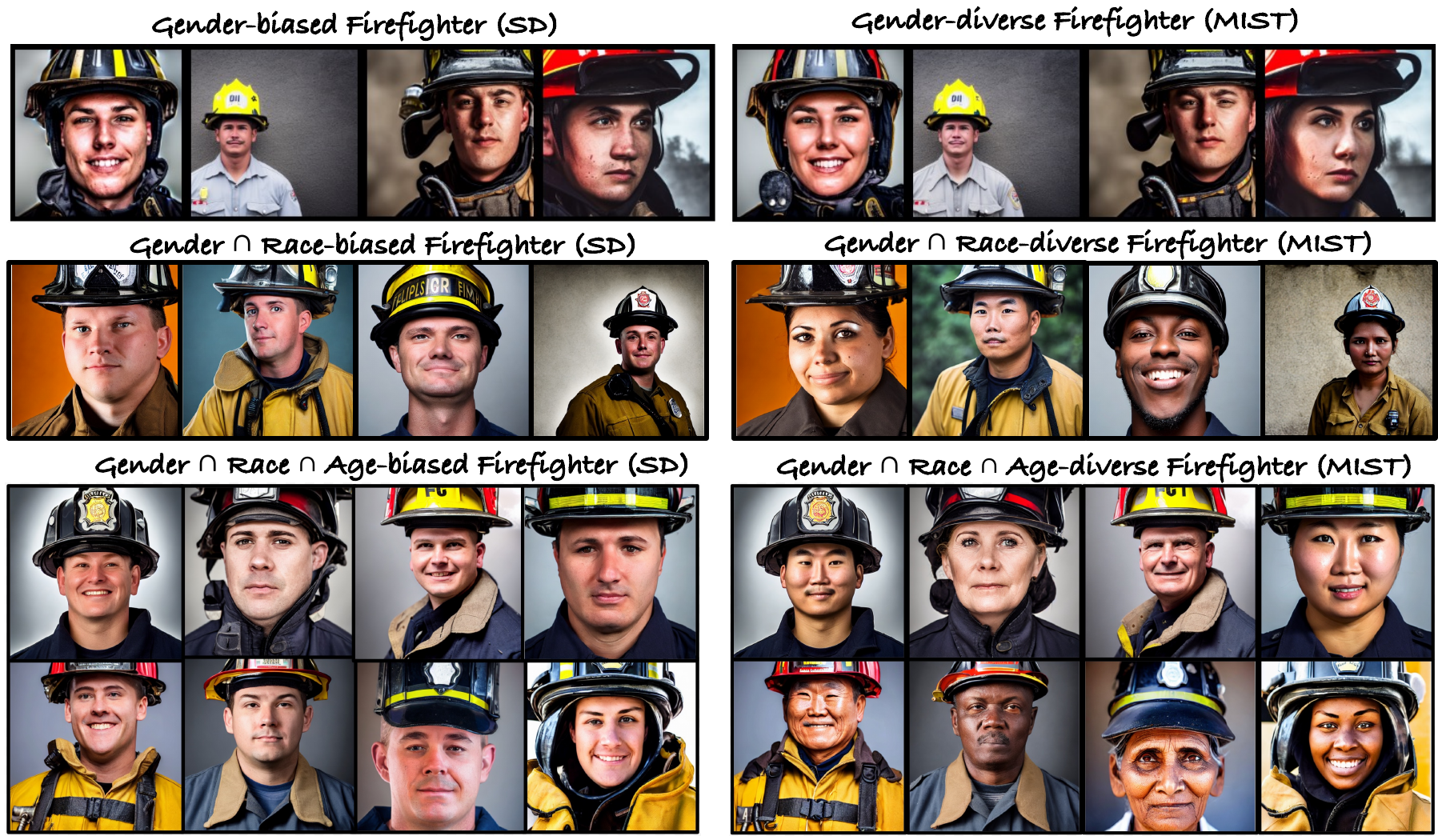}}
\\

\vspace{1em}
\end{tabular}
    \vspace{-2.5em}
    \captionof{figure}{Existing text-to-image model such as Stable Diffusion (SD) \cite{rombach2022high}   exhibit significant biases, including intersectional bias that affects people who are part of two or more marginalized groups (left). MIST finetunes the cross-attention maps of the SD model to mitigate biases related to single or intersectional attributes, such as (\textit{gender}), (\textit{gender  \& race \& age}) (right).} 
    \label{fig:teaser}
\end{center}
}]

%% file: sec/0_abstract.tex
\begin{abstract}
Diffusion-based text-to-image models have rapidly gained popularity for their ability to generate detailed and realistic images from textual descriptions. However, these models often reflect the biases present in their training data, especially impacting marginalized groups. While prior efforts to debias language models have focused on addressing specific biases, such as racial or gender biases, efforts to tackle intersectional bias have been limited. Intersectional bias refers to the unique form of bias experienced by individuals at the intersection of multiple social identities. Addressing intersectional bias is crucial because it amplifies the negative effects of discrimination based on race, gender, and other identities. In this paper, we introduce a method that addresses intersectional bias in diffusion-based text-to-image models by modifying cross-attention maps in a disentangled manner. Our approach utilizes a pre-trained Stable Diffusion model, eliminates the need for an additional set of reference images, and preserves  the original quality for unaltered concepts. Comprehensive experiments demonstrate that our method surpasses existing approaches in mitigating both single and intersectional biases across various attributes. We make our source code and debiased models for various attributes available to encourage fairness in generative models and to support further research.
  
\end{abstract}

%% file: sec/1_intro.tex
\section{Introduction}
\label{sec:intro}

Recently, there has been notable progress in text-based content generation, with models such as Stable Diffusion \cite{rombach2022high} leading the way in generating high-quality, realistic images from text prompts. These breakthroughs have led to their widespread adoption across various applications. However, a critical issue with these models is their tendency to perpetuate existing biases.   Since they are heavily data-driven and rely on large-scale multimodal datasets \cite{schuhmann2021laion} typically scraped from the internet, they can inadvertently reflect and amplify biases present in the source data.  These biases, which can be cultural, racial, or gender-based, impact the fairness and ethical considerations of the content they generate.

There has been some research aimed at mitigating  biases in generative models \cite{zhang2023iti, orgad2023editing, gandikota2023unified, friedrich2023fair}. However, these efforts often fall short in addressing \textit{intersectional} biases.   Intersectional bias refers to the specific kind of bias experienced by individuals belonging to two or more marginalized groups, such as \textit{a black woman}.  This kind of bias is particularly concerning because it combines multiple forms of discrimination, such as those based on \textit{race} or \textit{gender}, leading to compounded negative impacts. Imagine a generative model trained to generate images of professionals in various fields. If the training data for this model is skewed, it might disproportionately represent certain groups \cite{seth2023dear, birhane2021multimodal}. For instance, it might more frequently generate images of \textit{men} when prompted to create pictures of \textit{engineers} or \textit{scientists}, reflecting \textit{gender} bias \cite{bolukbasi2016man}. However, intersectional bias becomes evident when the model not only reflects \textit{gender} bias but also \textit{racial} bias. For example, when asked to generate images of \textit{women in leadership roles}, it might predominantly show \textit{white women}, thereby underrepresenting \textit{women of color}. This kind of bias reflects the intersection of \textit{gender} and \textit{racial} biases. Women of color are at a unique intersection of identities that are often underrepresented or misrepresented in data sets, leading to generative models that fail to accurately or fairly represent them. This is a critical issue because it perpetuates stereotypes and disregards the diverse experiences of individuals belonging to multiple marginalized groups.

One of the primary reasons for the biases in these models is the dataset on which they are trained. One approach to ensuring the fairness of generative models involves curating the training data to exclude any biased content, as suggested by previous studies \cite{saunders2020neural}. However, training large models is a costly endeavor, and the effects of data curation on a model can  have unpredictable and undesirable effects \cite{belkhale2023data}. 

More recent work \cite{orgad2023editing, gandikota2023unified} focused on edit implicit assumptions and biases in pre-trained diffusion models by updating the cross-attention layers. However, they might  inadvertently influence surrounding concepts when modifying a specific concept \cite{orgad2023editing}. For instance, trying to edit the concept of engineers to appear female might also inadvertently influence the representation of teachers, a profession often dominated by females, making it skew even more towards female representation. A recent work   \cite{gandikota2023unified} attempts to solve this drawback by taking a preservation set   into account.   However, they require  a manually curated list of concepts to preserve in order to prevent the model to influence surrounding concepts and face interdependencies that result in compounded biases, ultimately hindering their ability to address intersectional bias effectively.
 
In contrast, we propose \texttt{MIST} that modifies the cross-attention maps in  a \textit{disentangled} way. Our method exploits the structured nature of text embeddings in text-to-image diffusion models, adjusting the cross-attention layers to change the model's behavior without the need for retraining. This process involves altering only specific portions of the model's weights to address and reduce bias related to certain concepts. By adjusting these weights in a disentangled way, our approach is inherently capable of simultaneously tackling multiple biases—such as those associated with gender, race, and age—without affecting related concepts (see Fig. \ref{fig:teaser}). Our contributions are as follows:

\begin{itemize}
\item We introduce a novel approach that finetunes the cross-attention weights in text-to-image diffusion models in a disentangled manner without the necessity for re-training.
\item Our method stands out by its capability to simultaneously mitigate for biases in multiple attributes, effectively tackling intersectional bias issues. Through both qualitative and quantitative evaluations, our method demonstrates superior performance over previous methods in addressing both singular and multiple attribute biases.
\item Our method inherently designed to preserve surrounding attributes while eliminating biases, avoiding the need for manually curated lists or supplementary reference images. This efficiency and streamlined nature mark our approach as both effective and practical.
\item  Drawing inspiration from NLP, our technique capitalizes on the structured composition of text embeddings in text-to-image diffusion models to achieve disentangled editing. This not only paves the way for bias mitigation but also opens avenues for controlled image or video editing capabilities. 
\item We publicly share our source code and debiased models across a range of singular and intersectional attributes, aiming to advance the development of fair generative models.
\end{itemize} 
 
We have named our method \texttt{MIST}, inspired by the analogy of navigating through and clearing a misty landscape. This metaphor reflects our approach to effectively addressing intersectional biases in generative models, akin to the gradual process of unveiling and resolving hidden biases — similar to how one clears mist to reveal the true landscape. In doing so, \texttt{MIST} preserves the inherent generative quality of unaltered concepts, ensuring their balanced and accurate representation within the diffusion models. \\

\textbf{Disclaimer}: 
This paper contains representations of multiple biases and stereotypes. We want to emphasize that the primary objective of this research is to mitigate the biases that are already present in generative models, and we do not intend to discriminate against any identity groups or cultures in any way. 
We also recognize that gender is a spectrum, not a binary concept, and that it shouldn't be assumed based on appearance. However, given the constraints of our classification tools and to be able to make fair comparisons with previous studies, our discussion in this work will focus on the perceived binary gender.
 

%% file: sec/2_related.tex
\section{Related work}
\label{sec:related}
\paragraph{Bias mitigation.}
The study of fairness in generative models is an actively studied and rapidly developing research area \cite{karakas2022fairstyle, xu2018fairgan, sattigeri2019fairness, d2023improving, shen2023finetuning}. While research specifically targeting diffusion models is limited, there are several significant works. Earlier efforts have addressed this concern by adjusting model parameters, employing a methodology involving the projection of biased directions in the text embedding space such as Debias-VL \cite{chuang2023debiasing}, or defining projections over textual representations such as Concept Algebra \cite{wang2023concept}. TIME~\cite{orgad2023editing} proposes to fine-tune projection matrices in cross-attention layers to edit implicit assumptions in the text-to-image diffusion models. Authors project source prompts, which the model has assumptions about, to destination prompts thereby editing the model to reflect the desired outputs indicated by the target prompts. However, they edit the linear projections within cross-attention layer for each concept separately and their approach suffers from the fact that performing debiasing for specific attribute affects others,  which may be undesired. Unified Concept Editing (UCE) ~\cite{gandikota2023unified} extends TIME by enabling simultaneous edits on multiple concepts while preserving a predefined set of concepts that are not intended to change.  However, in order to maintain protected attributes, a need arises for the creation of specifically crafted manual lists, wherein the attributes deemed crucial for preservation are explicitly identified and listed. ITI-GEN \cite{zhang2023iti}  addresses the issue of fairness by training q-learnable tokens for every category $a_k^m$, such as \textit{male} and \textit{female}, related to the attribute $\mathcal{A}_m$, such as \textit{perceived gender}, with the objective of mitigating bias in the concerned attribute. . These learned tokens $S^m_k$ are then inserted into Inclusive Prompt Set $\mathcal{P}_{total}$  to form attribute-based learned token set. In inference time, an implicit token is randomly selected from $\mathcal{P}_{total}$ and injected into the original prompt $\mathbf{T}$ to obtain $P_k^m = [\mathbf{T};S_k^m]$. However, in order to learn implicit tokens, one must have an access to dozens of separate reference images for each category $a^m_k$ which ends up being a considerable limitation for their applicability in practice. 

\paragraph{Intersectional bias.} While most debiasing methods can be easily applied to multi-category debiasing, addressing intersectional bias presents unique challenges. For instance, debiasing based on gender may unintentionally favor one racial group over another, and vice versa. While some research has explored intersectional bias in NLP~\cite{lalor-etal-2022-benchmarking, lepori-2020-unequal, subramanian-etal-2021-evaluating, hassan-etal-2021-unpacking-interdependent} and GANs \cite{karakas2022fairstyle}, the investigation and mitigation of such biases in text-to-image diffusion models remain relatively unexplored.

\paragraph{Text-to-image diffusion models.} The development of large-scale text-to-image diffusion models~\cite{rombach2022high, nichol2021glide} has enabled the wide-spread use of image generation models \cite{chen2020uniter, kim2021vilt, saharia2022photorealistic} due to the abundance of image-text pair datasets~\cite{schuhmann2021laion, schuhmann2022laion} and their ease of training compared to Generative Adversarial Networks~\cite{goodfellow2014generative}. Despite their ability to create diverse and realistic images, these models often reflect and even amplify the biases present in the datasets used for training. In parallel to mentioned works, concept editing has been recently explored to change a model's behavior based on user instruction by editing a subset of model weights. This method has been applied to diffusion models for different purposes, including editing images based on text descriptions~\cite{kawar2023imagic,hertz2022prompt},personalization~\cite{ruiz2022dreambooth,gal2022textual}, suppressing inappropriate contents such as nudity and violence~\cite{schramowski2022safe, gandikota2023erasing, kumari2023ablating}.

%% file: sec/3_background.tex
\section{Background}
\label{sec:background}
\begin{figure*}[t]
  \includegraphics[width=\textwidth]{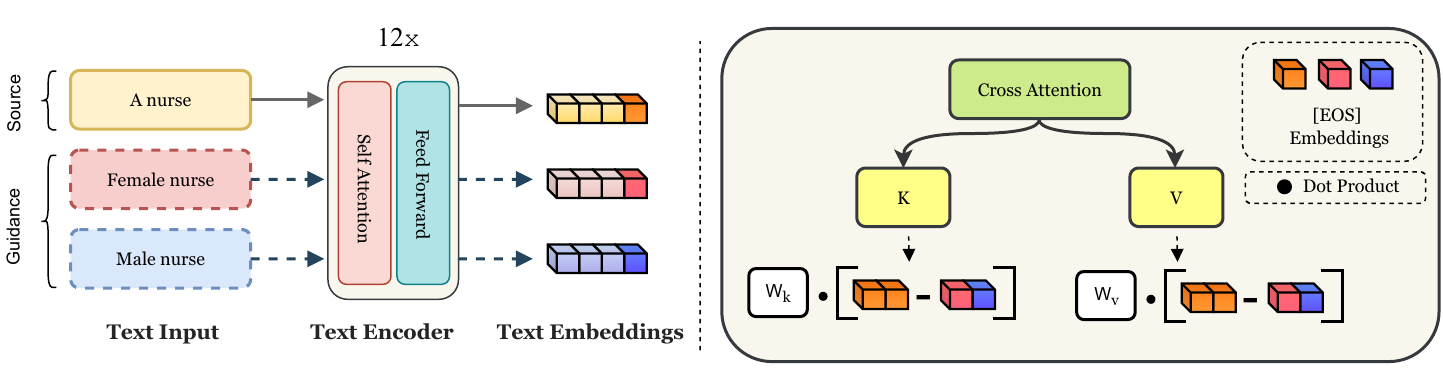}
  \caption{\textbf{Overview of the proposed method.} Given a source embedding $\mathcal{C}_s$ such as \textit{'A nurse'} and a guidance embedding $\mathcal{C}_g$ such as \textit{'A female nurse'}, \texttt{MIST} debiases the source attribute with respect to the guidance. In particular, we inject the \texttt{<EOS>} token from the guidance into the source embedding (left) to update the cross-attention layers in a disentangled manner (right). }
\end{figure*}
 
\paragraph{Diffusion models.} Diffusion models are a class of generative models that estimates the complex data distribution through iterative denoising process. As the source of diversity, the initial latent $x_T \sim \mathcal{N}(\textbf{0}, \textbf{I})$ is sampled and fed to U-Net \cite{ronneberger2015u} model, $\epsilon_\theta$, as an attempt to gradually denoise the latent variable $x_t$ to obtain $x_{t-1}$ for \textit{T} timesteps where $x_1$ corresponds to the final image. The joint probability of latent variables $\{x_1, ..., x_T\}$ is modeled as a Markov Chain.
\begin{equation}
    p_\theta(x_{1:T}) = p(x_T)\prod_{t=T}^{1}p_\theta(x_{t-1}|x_t)
\end{equation}
In text-to-image generation task, diffusion models such as Stable Diffusion \cite{rombach2022high} are conditioned on an external text input $c$ where the overall aim is producing an image that is aligned with the description provided by $c$. To train text-to-image models, diffusion models use a simplified objective.
\begin{equation}
    \mathbb{E}_{x, c, \epsilon, t}\left[\lVert \epsilon_\theta(x_t, t, c) - \epsilon \rVert_2^2\right],
\end{equation}
where $(x_t, c)$ is latent-text condition pair, $\epsilon \sim \mathcal{N}(\textbf{0}, \textbf{I})$ and $t \sim \mathcal{U}([0, 1])$. For unconditional generation, $c$ is set to null text. In inference stage, classifier free guidance is applied to noise prediction to improve the sample quality.
\begin{equation}
    \Tilde{\epsilon}_\theta(x_t, t, c) = \epsilon_\theta(x_t, t, \emptyset) + \gamma[\epsilon_\theta(x_t, t, c) - \epsilon_\theta(x_t, t, \emptyset)],
\end{equation}
where $\gamma \geq 1$ is the guidance scale and $\emptyset$ denotes the null text condition.

\paragraph{CLIP text encoder.} Stable Diffusion  employs the CLIP \cite{radford2021learning} text encoder for the purpose of transforming raw textual conditions   into encoded text embeddings \textit{C} = $\{c_{\text{[SOS]}}, c_1, ..., c_N, c_{\text{[EOS]}}\}$ where $c_i \in \mathbb{R}^{d \times 1}$.

\paragraph{Cross-attention.} The cross-attention mechanism serves as the bridge between textual and visual information, playing a pivotal role in associating visual significance with text tokens. It produces the linear projections Q, K and V where Q encodes the visual features by projection matrix $W_q$ while K and V encodes the information coming from the text-guidance by projection matrices $W_k, W_v$, respectively.   More formally, for any text embedding $c_i$ corresponding lower dimensional Key and Value embeddings are calculated as: $k_i = W_k \cdot c_i$ and $v_i = W_v \cdot c_i$ where $W_k, W_v \in \mathbb{R}^{m \times d}$ and $k_i, v_i \in \mathbb{R}^m$. Then, in parallel, K and V projections can be calculated as:
$K = W_k \cdot [c_{\text{[SOS]}}\hspace{4pt}c_1\hspace{4pt}...\hspace{4pt}c_{\text{[EOS]}}]$ and $V = W_v \cdot [c_{\text{[SOS]}}\hspace{4pt}c_1\hspace{4pt}...\hspace{4pt}c_{\text{[EOS]}}]$. Attention maps indicate relevance between each visual-text token pair, calculated as: 
\begin{equation}
    \mathcal{A} = \text{Softmax}\left(\frac{Q^TK}{\sqrt{m}}\right)
\end{equation}
Final output of the cross-attention layer is given as a linear combination of V with attention weights $\mathcal{A}$, calculated as $    \mathcal{O} = \mathcal{AV^T}$. The resultant activations then proceed through the consecutive layers of U-Net.

%% file: sec/4_methodology.tex
\section{Methodology}
\label{sec:methodology}

  Prior work \cite{orgad2023editing, gandikota2023unified} showed that it is possible to edit implicit assumptions and biases in pre-trained diffusion models by updating the cross-attention layers.  Let $\mathcal{C}_s$  be the source embedding, derived from the tokens of the source prompt such as \textit{`a CEO'}, and $\mathcal{C}_g$ be the corresponding destination embeddings such as \textit{`a female CEO'}. Then TIME \cite{orgad2023editing} updates projection matrices $W_k$ and $W_v$ to bring source embedding closer to the destination embedding, so that the model no longer makes the implicit assumption that CEOs are male. In more detail,  they minimize the following objective function: 

\begin{align}
    \min_{W} \sum_{i=0}^{m}||Wc_i -  v_i^*||_2^2 +\lambda ||W - W^\mathrm{old}||_F^2
    \label{eq:time}
\end{align}

\noindent where $c_i$ and $c_i^*$ are the corresponding source and destination embeddings, $W^\mathrm{old}$ is a given pre-trained projection layer, and  $v_i^* = W^\mathrm{old}c_i^*$.    However, this approach can inadvertently influence surrounding concepts when modifying a specific concept, as it affects all vector representations from  $c_1$ to $c_m$.  For instance, trying to edit the concept of CEOs to appear female might also inadvertently influence the representation of teachers, a profession often dominated by females, making it skew even more towards female representation. A recent work UCE \cite{gandikota2023unified} attempts to solve this drawback by taking a preservation set $P$ into account. Given a set of editing concepts $c_i \in E$, UCE maps to target values $v_i^* = W^\mathrm{old}_v c_{i^*}$ instead of the   $W^\mathrm{old}c_i$ in Eq.\ref{eq:time}. The idea is to  preserve outputs corresponding to the inputs $c_j\in P$ as $W^\mathrm{old}c_j$. More formally, 

\begin{align}
\min_{W} \sum_{c_i\in E}||Wc_i - v_i^*||_2^2 + \sum_{c_j\in P}||Wc_j - W^\mathrm{old}c_j||_2^2
\label{eq:uce}
\end{align}
 
However, this approach requires  a manually curated list of concepts to preserve in order to prevent the model to influence surrounding concepts.  For instance, in the process of debiasing a specific profession like \textit{firefighter}, a catalogue of $225$ professions\footnote{\url{https://github.com/rohitgandikota/unified-concept-editing/blob/main/data/profession_prompts.csv}} is given to ensure preservation which is challenging to curate manually.  Furthermore, as highlighted in \cite{gandikota2023unified}, when their method debiases in isolation it can perpetuate biases along other dimensions. Additionally, when addressing multiple attributes simultaneously, they report interdependencies that result in compounded biases, ultimately hindering their ability to address intersectional bias effectively.

 \begin{figure}[t]
  \centering
  \includegraphics[width=\linewidth]{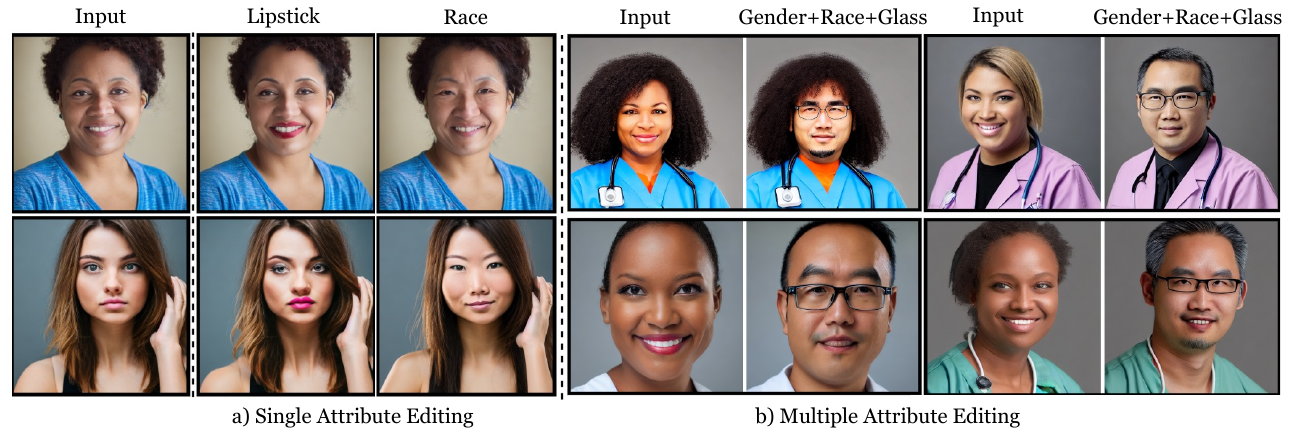} 
  \caption{\textbf{Observation: \texttt{<EOS>} token enables disentangled editing}: Given a text embedding $\mathcal{C}$ = $\{c_{\text{\texttt{<SOS>}}}, c_1, ..., c_N, c_{\text{\texttt{<EOS>}}}\}$, we observe that the generation process can be controlled in a highly disentangled manner by the end-of-sentence token $c_{\text{\texttt{<EOS>}}}$. (a) Original images are generated with prompt \textit{a woman} with \textit{lipstick} and \textit{race} edits applied. (b)   \texttt{<EOS>} can also handle edits involving multiple attributes simultaneously such as transforming input images with \textit{Asian male person with eyeglasses} prompt.}  
  \label{fig:intuition}
\end{figure}

In contrast, we propose to modify the    cross-attention maps in a  \textit{disentangled} way, inspired by a foundational observation. Given a text embedding $\mathcal{C} = \{c_{\text{\texttt{<SOS>}}}, c_1, ..., c_N, c_{\text{\texttt{<EOS>}}}\}$, we observe that the generation process can be controlled in a highly disentangled manner by the end-of-sentence token $c_{\text{\texttt{<EOS>}}}$ only. Fig. \ref{fig:intuition} illustrates this observation.  Starting with a source prompt $\mathcal{C}_s$, such as \textit{`a woman'}, and a guidance prompt $\mathcal{C}_g$, such as \textit{`a woman with lipstick'}, our method enables targeted edits where only the specified attribute (\textit{lipstick}, in this example) is modified, while the rest of the face remains faithful to the original image, as shown in Fig.~\ref{fig:intuition} (a). We further explored the capability of this intuition to handle edits involving multiple attributes simultaneously, an essential factor for achieving intersectional debiasing. Figure \ref{fig:intuition} (b) showcases the ability of our method to concurrently edit Gender, Race, and Eyeglass attributes  while preserving the original structure of the image. Specifically, the edits for each image are conducted by utilizing the  \texttt{\texttt{<EOS>}} token from the prompt \textit{`Asian male person with eyeglasses'}.  We attribute this effect to the fact that the CLIP text encoder is a unidirectional decoder-only transformer, hence it consolidates the activations from the highest transformer layer into the \texttt{<EOS>} token's embedding.   In fact, as a result of this causal language representation, \texttt{<EOS>} token frequently used a global feature representation of the entire text  in NLP tasks  \cite{chen2023mclip}.  We hypothesize that a comparable mechanism is at play in image generation and editing, with the \texttt{<EOS>} token capturing the global context during the image creation process. To the best of our knowledge, we are the first to exploit this insight for the purpose of disentangled image editing. This discovery led us to the realization that the \texttt{<EOS>} token could facilitate targeted  manipulation of the cross-attention maps in a disentangled manner. Therefore, drawing inspiration from this observation, we can formulate an  optimization problem aimed at fine-tuning the parameters associated with key and value matrices such that we mitigate inherent biases present in $\mathcal{G}_s$ by taking guidance from global context of $\mathcal{G}_g$. Formally, given the guidance  context $\mathcal{G}_g$ and a source prompt $\mathcal{G}_s$  to debias, we  formulate the debiasing problem as a constrained minimization problem as follows:

\begin{align}
\label{eq:obj}
\min_{W^*} ||W^*c_{g_{\text{\texttt{<EOS>}}}} - W^*c_{s_{\text{\texttt{<EOS>}}}}||_2^2 + \lambda ||W^* - W^{old}||_2^2 
\end{align}

\noindent where $W^*$ is the finetuned projection matrix, $c_{g_{\text{\texttt{<EOS>}}}}$ is the $\text{\texttt{<EOS>}}$ token of the guidance concept, $c_{s_{\text{\texttt{<EOS>}}}}$ is the \text{\texttt{<EOS>}} token of the source concept. $\lambda$ is   a regularization hyper-parameter. 

 \begin{figure*} 
 \centering
    \includegraphics[width=0.85\linewidth]{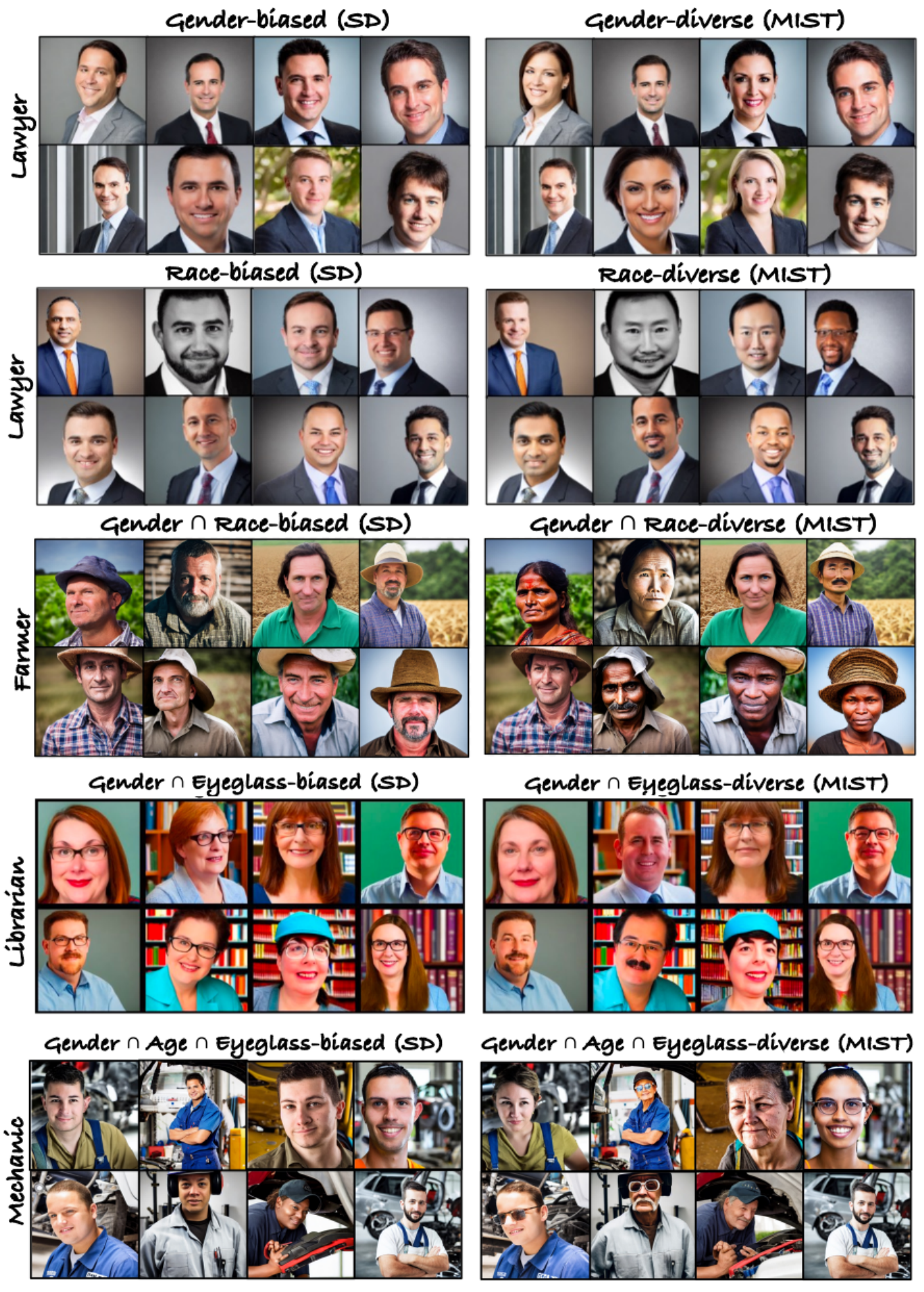}   
\caption{\textbf{Qualitative results on singular and intersectional debiasing}. Samples generated with the same seed using Stable Diffusion are displayed on the left, while samples produced with \texttt{MIST} are shown on the right. Our approach effectively  debiases single attributes like gender and race, as well as intersectional attributes such as \textit{Race \& Gender}, and triple attributes like \textit{Gender \& Age \& Eyeglasses}. }
\hfill
\label{fig:qualitative_single}
\end{figure*}

\paragraph{Intersectionality.} A key benefit of our debiasing approach compared to prior methods is its ability to address biases across multiple categories simultaneously with just a single token, eliminating the need to process each token for every biased category individually. This approach effectively removes the summation component found in previous works (refer to Eq. \ref{eq:time} and Eq. \ref{eq:uce}). Moreover, unlike \cite{gandikota2023unified} our method does not necessitate the use of a reference image set for each attribute of interest.  We can generalize our formulation to cover intersectional biases.  Let us define $\Delta_{\text{\texttt{<EOS>}}}$:

\begin{equation}
    \Delta_{\text{\texttt{<EOS>}}} = [c_{g_{\text{\texttt{<EOS>}}}} - c_{s_{\text{\texttt{<EOS>}}}}]
\end{equation}

Given L guidances for each intersectional attribute, we formulate intersectional bias mitigation as:  
\begin{equation}
\begin{aligned}
    \min_{W^*}  & \bigg\lVert W^*
    \begin{bmatrix}
    \Delta^{(1)}_{\text{\texttt{<EOS>}}}  
    \hspace{2pt}\hdots\hspace{2pt} \Delta^{(L)}_{\text{\texttt{<EOS>}}}    
    \end{bmatrix} 
     \bigg\rVert_F^2 + \lambda ||W^* - W^{old}||_2^2\\\\  
\end{aligned}
\label{eq:intersection}
\end{equation}

Given that our method can perform disentangled edits,  we can address intersectional biases in a single debiasing process instead of resorting to an iterative approach. Given Eq. \ref{eq:intersection}, we fine-tune projection matrices in cross-attention layers until a uniform balance between attributes is achieved. For example, when debiasing a singular attribute like gender within a particular profession such as Lawyer, and given $N$ randomly sampled images, we aim for an equal distribution, with 'A female lawyer' representing 50\% (similarly, for 'A male lawyer'). To assess the balance of specific attributes for debiasing, we utilize CLIP.

%% file: sec/5_experiments.tex
\section{Experiments}
\label{sec:experiments}

\begin{table*}[t!] 
  \begin{minipage}{1\textwidth} 
    \label{tab:bias_comparison}
    \centering
    \begin{tabular}{|l|c|c|c|c | c|c|}
      \toprule  \toprule
      \multirow{2}{*}{Occupation} & \multicolumn{1}{c}{\textbf{SD}} & \multicolumn{1}{c}{\textbf{MIST}} & \multicolumn{1}{c}{\textbf{UCE}} & \multicolumn{1}{c}{\textbf{TIME}} & \multicolumn{1}{c}{\textbf{Concept Algebra}} & \multicolumn{1}{c}{\textbf{Debias-VL}}\\
       \cmidrule(lr){2-2} \cmidrule(lr){3-3}  \cmidrule(lr){4-4} \cmidrule(lr){5-5} \cmidrule(lr){6-6} \cmidrule(lr){7-7}
        \quad $\psi (\downarrow)$ & $0.86 \pm 0.22$ &  $0.14 \pm 0.08$ & $0.28 \pm 0.16$ & $0.38 \pm 0.11$ & $0.63 \pm 0.08$   & $0.60 \pm 0.04$ \\
      \midrule
      \midrule 
      Teacher & 0.36 & \textbf{0.04} & 0.06 & 0.34 & 0.46 & 0.11\\ 
      Nurse & 1.00 & \textbf{0.26} & 0.39 & 0.34  & 0.91 & 0.87\\ 
      Librarian & 0.92 & \textbf{0.04} & 0.07 & 0.26 & 0.66 & 0.34\\ 
      Hairdresser & 0.96  & 0.24 & \textbf{0.16} & 0.32 & 0.37 & 0.61\\ 
      Housekeeper & 1.00  & \textbf{0.10} & 0.41 & 0.32  & 0.68 & 0.80\\ 
      Developer & 0.98 & \textbf{0.14} &  0.51 & 0.50 & 0.74 & 0.90\\ 
      Farmer & 0.96 & \textbf{0.18} & 0.41 & 0.46 & 0.58& 0.97\\ 
      CEO & 0.96 & \textbf{0.20} & 0.28 & 0.28 & 0.25& 0.37\\ 
      Doctor & 0.64 & \textbf{0.12} & 0.20 & 0.58 & 0.40& 0.50\\ 
      \bottomrule
    \end{tabular}
    \caption{\textbf{Quantitative comparison on biasedness score}. Biasedness score $\psi$ of debiased concepts randomly selected from WinoBias dataset. $\psi = 0$ means perfect debiasing.   \texttt{MIST}   surpasses competing approaches and achieving lower scores across the majority of professions.}
    \label{table:biasedness_1}
  \end{minipage}%
\end{table*}

In this section, we conduct qualitative and quantitative experiments to evaluate the efficiency of our approach.  Please see Supplementary Material for more qualitative and quantitative comparisons.

\subsection{Experimental Setup.} 

\paragraph{Baselines.}  
We compare our approach  TIME \cite{orgad2023editing}, UCE \cite{gandikota2023unified}, Concept Algebra   \cite{wang2023concept}, and Debias-VL  \cite{chuang2023debiasing} (please see Supplementary Material for a comparison against ITI-GEN \cite{zhang2023iti}).   As discussed before, TIME and UCE edits cross attention maps to perform debiasing.    However, TIME inadvertently influences surrounding concepts when modifying a specific concept. On the other hand, UCE   requires  a manually  curated list of concepts to preserve.  Debias-VL  modifies model parameters after training, by projecting out biased directions in the text embedding, while Concept Algebra performs algebraic manipulation of the representations.

\paragraph{Dataset.} Stable Diffusion displays biases related to gender and race, among other attributes, when generating images for professional titles.  For example, when prompted to generate "an image of a CEO," the model produces images depicting females only 6\% of the time.  Moreover, the model exhibits biases towards other attributes, such as eyeglasses; when asked to generate "an image of a CEO," Stable Diffusion depicts males with eyeglasses in 88\% of the images. This behavior amplifies the bias observed in the LAION dataset, where searching for "a photo of the face of a CEO" results in 77\% of the images from a random selection of 1,000 showing males with eyeglasses. Consequently, in alignment with prior research, we adopt the WinoBias \cite{zhao2018gender} dataset, which encompasses 36 professions (see Supplementary Material  for more details).

\paragraph{Implementation details. }
We conducted our experiments using Stable Diffusion 1.5 and utilized the official repositories of baseline methods. Our experiments were performed on a single L40 GPU. In Eq. \ref{eq:intersection}, we set the $\lambda$ term as $\lambda = 1/L$, where $L$ represents the number of concepts undergoing debiasing. In each iteration, we computed the CLIP score on $N=200$ images to evaluate whether a uniform distribution across attributes was achieved.  For optimization, we used a learning rate of $0.5$.

\begin{table*}[t] 

  \begin{minipage}{1\textwidth} 
      \centering
    \begin{tabular}{| l| c|c|c|c|c }
      \toprule 
      \multirow{2}{*}{Occupation} & \multicolumn{1}{c}{\textbf{MIST (Ours)  vs SD}} & \multicolumn{1}{c}{\textbf{TIME  vs SD}} & \multicolumn{1}{c}{\textbf{Concept Algebra  vs SD}} & \multicolumn{1}{c}{\textbf{UCE  vs SD}}   \\
      \cmidrule(lr){2-2} \cmidrule(lr){3-3}  \cmidrule(lr){4-4} \cmidrule(lr){5-5}
      & Pixel Shift $(\downarrow)$ &   Pixel Shift  $(\downarrow)$ &  Pixel Shift $(\downarrow)$ &  Pixel Shift $(\downarrow)$ \\
      \midrule 
      Teacher & \textbf{97.71 $\pm$ 63.57} & 127.47 $\pm$  57.91& 143.96 $\pm$ 53.38 & 146.77 $\pm$ 59.29\\
      Nurse & \textbf{86.71 $\pm$ 55.31} & 126.35 $\pm$ 45.90& 142.72 $\pm$ 42.77& 114.84 $\pm$ 43.36 \\
      Librarian & \textbf{88.26 $\pm$ 53.30} & 149.62 $\pm$ 58.28 & 143.55 $\pm$ 50.07 & 92.40 $\pm$ 46.94\\
      Hairdresser & \textbf{81.78 $\pm$ 47.91} &118.57 $\pm$ 48.73 & 162.02 $\pm$ 41.23&  83.67 $\pm$ 41.03\\
      Housekeeper & \textbf{80.57 $\pm$ 55.04} &106.37 $\pm$ 53.41 & 138.24 $\pm$ 37.62& 83.76 $\pm$ 44.12 \\
      Farmer & 94.73 $\pm$ 39.93& 102.20 $\pm$ 40.31& 108.54 $\pm$ 47.80 & \textbf{90.94 $\pm$ 36.39}\\
      Sheriff & \textbf{94.53 $\pm$ 48.48} & 145.67 $\pm$ 43.65 & 102.96 $\pm$ 41.92& 124.44 $\pm$ 50.38 \\
      CEO & \textbf{54.86 $\pm$ 45.53}& 111.62 $\pm$ 54.12 & 102.30 $\pm$ 42.21& 99.58 $\pm$ 51.25 \\
      Cashier & \textbf{85.25 $\pm$ 47.17} & 145.70 $\pm$ 54.94& 104.67 $\pm$ 43.16 & 125.43 $\pm$ 40.21\\
      \bottomrule
    \end{tabular}
    \caption{\textbf{Content Preservation.}   We report  the average pixel-wise difference, calculated across 250 images generated for each non-debiased occupation. \texttt{MIST} consistently maintains a lower score across the majority of occupations, demonstrating its ability to preserve surrounding concepts while debiasing the desired attribute. }
    \label{tab:cvpr_comparison}
  \end{minipage}%
\end{table*}

\paragraph{Evaluation metrics.} For quantitative experiments, we report \textbf{biasedness metric},  \textbf{deviation of ratios} and \textbf{average pixel shift}.

\noindent \textbf{Biasedness metric $\psi$:} We calculate the deviation from the ideal ratio using the  biasedness metric to measure bias following~\cite{orgad2023editing}:

\begin{equation}
\psi = |r_{\text{ideal}}-r_{\text{actual}}|/{r_{\text{ideal}}}
\end{equation}

Biasedness metric  is $\psi=0$ when the model is completely debiased, and $\psi=1$ when generative process is completely biased toward one category within a specific attribute. $r_{\text{ideal}}=1/L$ where $L$ is the number of attributes to be debiased. For instance, if we are performing debiasing only on Gender, then  $r_{\text{ideal}}=0.5$.

\noindent \textbf{Deviation of ratios $\xi$:} When debiasing on a specific concept, it is crucial to preserve other concepts to prevent unintentional bias amplification. Given a concept set P that is not intended to change original distribution it belongs to, following \cite{gandikota2023unified} we measure the deviation of ratios as: 
\begin{equation}
 \xi =  \dfrac{1}{|P|}\sum_{p\epsilon P} |\delta^{(p)}_{\text{attribute}}-\delta^{(p)}_{\text{SD}}|
\end{equation} 
where $\delta = (r_{\text{ideal}}-r_{\text{actual}}) / r_{\text{ideal}}$. It differs from $\psi$  by avoiding the absolute value. This distinction is significant because a change in the bias direction, say from male to female, should yield a larger deviation.  

\noindent \textbf{Average pixel shift (APS):} Average pixel shift   for N samples is defined as: 
\begin{equation}
    \frac{1}{N} \sum_{i=1}^{N} \lVert \mathcal{I}_{\text{SD}}^{(i)} - \mathcal{I}_{\text{method}}^{(i)} \rVert_2, 
\end{equation}
where $\mathcal{I}_{\text{SD}}$ refers to images sampled from Stable Diffusion and $\mathcal{I}_{\text{method}}$ refers to images sampled from specific method being applied.

\subsection{Single attribute bias mitigation}
\label{sec:single}
Given the absence of alternative methods specifically aimed at mitigating intersectional bias in diffusion-based text-to-image models, we first present both qualitative and quantitative outcomes for bias reduction concerning singular attributes such as \textit{gender} and \textit{race}. A qualitative comparison can be viewed from Fig. \ref{fig:teaser} and Fig. \ref{fig:qualitative_single}. As evident from the visuals, \texttt{MIST} can debias based on a singular attribute in a disentangled manner. As a quantitative experiment, we employ a debiasing strategy on 35 professions identified in the WinoBias dataset. The evaluation of debiasing effectiveness involves the generation of 1000 images for each profession, followed by the computation of the biasedness metric $\psi$ through CLIP classification. A quantitative analysis involving our approach, Stable Diffusion, UCE, TIME, Concept Algebra, and Debias-VL is provided in Table \ref{table:biasedness_1}, where we highlight the results for a randomly selected subset of 9 occupations from the WinoBias dataset (see Supplementary Material for the full list). Our method not only markedly reduces gender bias within Stable Diffusion but also surpasses the performance of other debiasing techniques in this context.

 \begin{figure*}[t]
 \centering
    \includegraphics[width=1\linewidth]{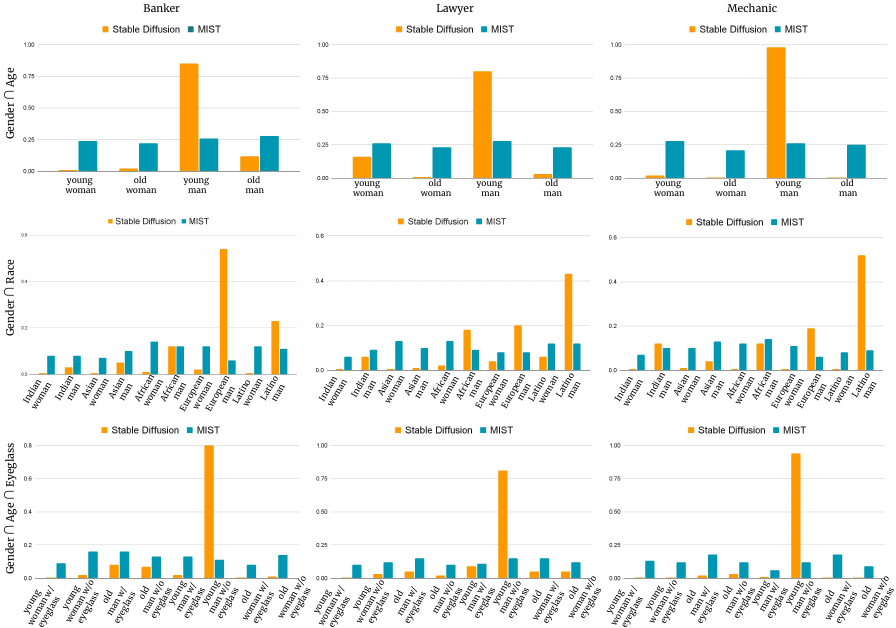}   
 \caption{\textbf{Quantitative comparison on intersectional bias}.  \texttt{MIST} achieves a uniform distribution across intersectional attributes in the majority of cases.  }
\hfill
\label{fig:intersectional_quantitative}
\end{figure*}

\subsection{Intersectional bias mitigation}

Fig. \ref{fig:teaser} and Fig. \ref{fig:qualitative_single} provide a qualitative demonstration of intersectional bias mitigation across two dimensions: gender \& race, gender \& eyeglasses, as well as across three dimensions: gender \& race \& age and  gender \& race \& eyeglasses. It's important to note that in this context, race is not binary but comprises six distinct categories: Indian, Asian, African, European, Latino, and Middle Eastern. As evident from the visuals, our method effectively debiases intersectional concepts in a disentangled manner (see Supplementary Material for more visuals).  Additionally, we offer a quantitative comparison between Stable Diffusion and \texttt{MIST} across various intersectional attributes (refer to Fig. \ref{fig:intersectional_quantitative}). Specifically, we present the ratio of each attribute (e.g., 'young women') using a CLIP classifier on $500$ randomly generated samples. \texttt{MIST} achieves a uniform distribution across intersectional attributes in the majority of cases. However, it's important to note that the CLIP classifier is known to have biases, particularly towards race \cite{gandikota2023unified}, so the results for Gender \& Race should be interpreted with caution.

\subsection{Preservation on remaining concepts}
When debiasing one concept, unintended changes in other related concepts can occur, potentially amplifying bias and affecting image fidelity. Our model has a strong capability of preserving concepts that are not intended to change due to its disentangled nature. In order to test the effectiveness of our method in preserving surrounding concepts, we   report the average pixel shift scores  (see Table \ref{tab:cvpr_comparison}).   In our experiments, we sample 250 images for each non-debiased occupation. The average pixel shift measures the change in each pixel between debiased image and the image generated from Stable Diffusion for the same random seed. We demonstrate that MIST exhibits the least shift among its competitors without requiring any hand-crafted preservation lists.

%% file: sec/6_conclusion.tex
\section{Limitations and Broader Impact}
Our method utilizes the pre-trained Stable Diffusion model, and as such, its image manipulation capabilities are significantly shaped by the datasets on which Stable Diffusion was trained and the language model CLIP that is known to have biases.  This becomes particularly relevant in the context of debiasing specific attributes, such as racial characteristics like \textit{black} or \textit{asian}. Utilizing the CLIP classifier to categorize race and gender introduces a range of potential issues, mainly due to inherent biases embedded within the classifier's training data. These biases can lead to misclassification and perpetuate stereotypes, reflecting and amplifying existing societal biases.  Moreover, the use of CLIP, despite its biases, is partly due to the lack of viable alternatives that offer the same level of performance and flexibility. Particularly, we use CLIP to bridge the gap between visual data and textual descriptions, making it particularly crucial for our task.   Nevertheless, it is important to recognize that the debiasing process itself may be influenced by biases that are already present in the model.

\section{Conclusion}
\label{sec:conclusion}
In this study, we introduce a novel method aimed at debiasing pre-trained text-to-image diffusion models. Our approach modifies the cross attention maps in a disentangled manner. Our comprehensive quantitative and qualitative analyses demonstrate that MIST significantly surpasses competing methodologies in performance. Notably, this is the first method designed to mitigate intersectional biases in text-to-image diffusion models, marking a pioneering step in the field. Furthermore, we emphasize the critical importance of tackling issues related to bias and fairness in diffusion models, underscoring that these considerations are essential for the development of ethical and fair AI technologies.
 

%% file: sec/7_appendix.tex
\clearpage
\makeatletter
\renewcommand \thesection{S.\@arabic\c@section}
\renewcommand\thetable{S.\@arabic\c@table}
\renewcommand \thefigure{S.\@arabic\c@figure}
\makeatother
\setcounter{section}{0}
\setcounter{page}{1}
\maketitlesupplementary

\section{Quantitative Comparison with ITI-GEN}

Due to space limitations, we were not able to provide a comparison against another debiasing method, ITI-GEN \cite{zhang2023iti} in the main paper. Unlike our method, ITI-GEN necessitates a reference image set for each attribute, thereby demanding manual effort for the debiasing process due to its need for assembling such image collections. We provide two quantitative comparisons using biasedness score (see Tab. \ref{table:biasedness}), and content preservation (see Tab. \ref{tab:preserve}).  

Tab. \ref{table:biasedness} presents the Biasedness metric results for a randomly chosen subset of 9 occupations from the WinoBias dataset. The Biasedness metric, denoted as $\psi$, equals $0$ when the model is fully debiased and $1$ when the generative process is entirely biased towards one category within a specific attribute. Our method significantly lowers gender bias within Stable Diffusion for the majority of the occupations compared to ITI-GEN.

Similarly, we compared both approaches in terms of content preservation. Debiasing one concept may inadvertently alter other related concepts, which could amplify bias or compromise image quality. To evaluate our method's proficiency in preserving surrounding concepts, we present the average pixel shift scores in Tab. \ref{tab:preserve}. The results indicate that  MIST exhibits the least shift compared to ITI-GEN without requiring any hand-crafted preservation lists.

\begin{table*}[t] 
  \begin{minipage}{1\textwidth} 
    \label{tab:bias_comparison}
    \centering
    \begin{tabular}{|l|c|c|c|c | c|c|}
      \toprule  
      \multirow{2}{*}{Occupation} & \multicolumn{1}{c}{\textbf{SD}} & \multicolumn{1}{c}{\textbf{MIST}} & \multicolumn{1}{c}{\textbf{ITI-GEN}}  \\
       \cmidrule(lr){2-2} \cmidrule(lr){3-3}  \cmidrule(lr){4-4} 
        \quad $\psi (\downarrow)$ & $0.86 \pm 0.22$ &  $0.14 \pm 0.08$ & $0.34 \pm 0.31$ \\
      
      \midrule
      
      \midrule 
      Teacher & 0.36 & \textbf{0.04} & 0.11 \\ 
      Nurse & 1.00 & \textbf{0.26} & 0.82 \\ 
      Librarian & 0.92 & \textbf{0.04} & 0.23\\  
      Hairdresser & 0.96  & 0.24 & \textbf{0.08} \\ 
      Housekeeper & 1.00  & \textbf{0.10} & 0.90 \\ 
      Developer & 0.98 & 0.14 &  \textbf{0.10} \\ 
      Farmer & 0.96 & \textbf{0.18} & 0.48 \\ 
      CEO & 0.96 & \textbf{0.20} & 0.26  \\ 
      Doctor & 0.64 & \textbf{0.12} & 0.16 \\ 
      \bottomrule
    \end{tabular}
    \caption{\textbf{Quantitative results on biasedness score  compared to ITI-GEN \cite{zhang2023iti}}. Biasedness score $\psi$ of debiased concepts randomly selected from WinoBias dataset. $\psi = 0$ means perfect debiasing. }
    \label{table:biasedness}
  \end{minipage}%
\end{table*}

\begin{table*}[] 
  \begin{minipage}{1\textwidth} 
      \centering
    \begin{tabular}{| l| c|c|c|c|c }
      \toprule 
      \multirow{2}{*}{Occupation} & \multicolumn{1}{c}{\textbf{MIST (Ours)  vs SD}} & \multicolumn{1}{c}{\textbf{ITI-GEN  vs SD}}   \\
      \cmidrule(lr){2-2} \cmidrule(lr){3-3}  
      & Pixel Shift $(\downarrow)$ &   Pixel Shift  $(\downarrow)$  \\
      \midrule 
      Teacher & \textbf{97.71 $\pm$ 63.57} & 114.35 $\pm$  26.59\\
      Nurse & \textbf{86.71 $\pm$ 55.31} & 118.49 $\pm$ 46.18\\
      Librarian & \textbf{88.26 $\pm$ 53.30} & 102.14 $\pm$ 37.32 \\
      Hairdresser & \textbf{81.78 $\pm$ 47.91} &98.44 $\pm$ 56.19 \\
      Housekeeper & \textbf{80.57 $\pm$ 55.04} &181.17 $\pm$ 61.19 \\
      Farmer & \textbf{94.73 $\pm$ 39.93}& 148.29 $\pm$ 47.10\\
      Sheriff & \textbf{94.53 $\pm$ 48.48} & 129.87 $\pm$ 39.45 \\
      CEO & \textbf{54.86 $\pm$ 45.53}& 151.72 $\pm$ 70.41 \\
      Cashier & \textbf{85.25 $\pm$ 47.17} & 137.66 $\pm$ 42.86\\
      Developer & \textbf{78.56 $\pm$ 41.31} & 193.09 $\pm$ 59.01 \\
      \bottomrule
    \end{tabular}
    \caption{\textbf{Content Preservation compared to ITI-GEN \cite{zhang2023iti}.} We report  the average pixel-wise difference, calculated across 250 images generated for each non-debiased occupation. \texttt{MIST} consistently maintains a lower score across the majority of occupations, demonstrating its ability to preserve surrounding concepts while debiasing the desired attribute. }
    \label{tab:preserve}
  \end{minipage}%
\end{table*}

\section{Additional Quantitative Results}
Table \ref{table:biasedness} presents a thorough comparison of several methods - SD,  TIME \cite{orgad2023editing}, UCE \cite{gandikota2023unified}, Concept Algebra (CAL) \cite{wang2023concept}, and Debias-VL  \cite{chuang2023debiasing} - against our proposed method, MIST, across a broad range of occupations from WinoBias dataset \cite{zhao2018gender}. The results indicate that our method surpasses its competitors in most of the occupations, highlighting the effectiveness of MIST.

\begin{table*}[t] 
  \begin{minipage}{1\textwidth} 
    \label{tab:bias_comparison}
    \centering
    \begin{tabular}{lcccc cc}
      \toprule \toprule
      \multirow{2}{*}{Occupation} & \multicolumn{1}{c}{\textbf{SD}} & \multicolumn{1}{c}{\textbf{UCE}} & \multicolumn{1}{c}{\textbf{MIST}} & \multicolumn{1}{c}{\textbf{TIME}} & \multicolumn{1}{c}{\textbf{CAL}} & \multicolumn{1}{c}{\textbf{Debias-VL}}\\
       \cmidrule(lr){2-2} \cmidrule(lr){3-3}  \cmidrule(lr){4-4} \cmidrule(lr){5-5} \cmidrule(lr){6-6} \cmidrule(lr){7-7}
        \quad $\psi (\downarrow)$ & $0.67 \pm 0.01$ &  $0.22 \pm 0.00$ & $0.13 \pm 0.00$ & $0.44 \pm 0.00$ & $0.43 \pm 0.01$   & $0.55 \pm 0.01$ \\
      
      \midrule \midrule 
      Attendant & 0.13 $\pm$ 0.06 & 0.09 $\pm$ 0.04 & \textbf{0.06} $\pm$ 0.04 & 0.50  $\pm$ 0.01& 0.23  $\pm$ 0.08 & 0.30 $\pm$ 0.04 \\ 
      Cashier & 0.67 $\pm$ 0.04&  0.16 $\pm$ 0.06& \textbf{0.04} $\pm$ 0.03 & 0.46   $\pm$ 0.01& 0.71  $\pm$ 0.10 & 0.23 $\pm$ 0.07\\ 
      Teacher & 0.42 $\pm$ 0.01& 0.06 $\pm$ 0.02 & \textbf{0.04} $\pm$ 0.05 & 0.34   $\pm$ 0.06& 0.46  $\pm$ 0.00 & 0.11 $\pm$ 0.05\\ 
      Nurse & 0.99  $\pm$ 0.01& 0.39$\pm$ 0.07 & \textbf{0.26} $\pm$ 0.01 & 0.34  $\pm$ 0.03& 0.91  $\pm$ 0.05& 0.87 $\pm$ 0.01\\ 
      Assistant & 0.19  $\pm$ 0.05& 0.14 $\pm$ 0.06 & \textbf{0.10} $\pm$ 0.02  & 0.32   $\pm$ 0.06& 0.20  $\pm$ 0.07& 0.35 $\pm$ 0.15\\ 
      Secretary & 0.88 $\pm$ 0.01&  0.10 $\pm$ 0.10&  \textbf{0.08} $\pm$ 0.10 & 0.58  $\pm$ 0.08& 0.65  $\pm$ 0.07& 0.65 $\pm$ 0.01\\ 
      Cleaner & 0.38 $\pm$ 0.04 &  0.33 $\pm$ 0.07 & \textbf{0.12} $\pm$ 0.03 & 0.58  $\pm$ 0.07& 0.11 $\pm$ 0.06& 0.18 $\pm$ 0.04\\ 
      Receptionist & 0.99 $\pm$ 0.01&  0.38 $\pm$ 0.01& \textbf{0.12} $\pm$ 0.05 & 0.36  $\pm$ 0.10& 0.90 $\pm$ 0.08& 0.74 $\pm$ 0.04\\ 
      Clerk & 0.10 $\pm$ 0.07&  0.23 $\pm$ 0.06 & \textbf{0.14} $\pm$ 0.04 & 0.58  $\pm$ 0.03& 0.11 $\pm$ 0.08& 0.10 $\pm$ 0.04\\ 
      Counselor & 0.06 $\pm$ 0.05& 0.40 $\pm$ 0.02  & \textbf{0.06} $\pm$ 0.08 & 0.74  $\pm$ 0.08& 0.30  $\pm$ 0.03& 0.10 $\pm$ 0.07\\ 
      Designer & 0.23 $\pm$ 0.05&  0.07 $\pm$ 0.05 & \textbf{0.04} $\pm$ 0.03 & 0.44  $\pm$ 0.06 & 0.25  $\pm$ 0.12& 0.48 $\pm$ 0.06\\ 
      Hairdresser & 0.74 $\pm$ 0.11 &  \textbf{0.16} $\pm$ 0.04 & 0.26 $\pm$ 0.04 & 0.32  $\pm$ 0.01& 0.37  $\pm$ 0.16& 0.61 $\pm$ 0.04\\ 
      Writer &  0.15 $\pm$ 0.03& 0.24 $\pm$ 0.08&  0.12 $\pm$ 0.03 & 0.54  $\pm$ 0.08& \textbf{0.07}  $\pm$ 0.03& 0.45 $\pm$ 0.04\\ 
      Housekeeper & 0.81 $\pm$ 0.04 &  0.41 $\pm$ 0.05& \textbf{0.10}  $\pm$ 0.07 & 0.32 $\pm$ 0.03 & 0.68  $\pm$ 0.18& 0.80 $\pm$ 0.07\\ 
      Baker &  0.81 $\pm$ 0.01&  0.29 $\pm$ 0.08& \textbf{0.08} $\pm$ 0.03 & 0.40  $\pm$ 0.04& 0.74  $\pm$ 0.04& 0.72 $\pm$ 0.05\\ 
      Librarian & 0.86 $\pm$ 0.06&  0.07 $\pm$ 0.07& \textbf{0.04} $\pm$ 0.06 & 0.26  $\pm$ 0.05& 0.66 $\pm$ 0.07& 0.34 $\pm$ 0.06\\ 
      Tailor & 0.30 $\pm$ 0.01&  \textbf{0.27} $\pm$ 0.01& 0.34 $\pm$ 0.07 & 0.50 $\pm$ 0.03& 0.21 $\pm$ 0.05& 0.33 $\pm$ 0.11\\ 
      Driver & 0.97 $\pm$ 0.02&  0.21 $\pm$ 0.07& \textbf{0.13} $\pm$ 0.01 & 0.48 $\pm$ 0.09& 0.20  $\pm$ 0.07& 0.65 $\pm$ 0.04\\ 
      Supervisor & 0.50 $\pm$ 0.01&  0.26 $\pm$ 0.04& 0.16 $\pm$ 0.02& 0.50$\pm$ 0.07 & \textbf{0.07}  $\pm$ 0.03& 0.43 $\pm$ 0.04\\ 
      Janitor & 0.91 $\pm$ 0.05&  0.16 $\pm$ 0.04& \textbf{0.12} $\pm$ 0.05 & 0.36  $\pm$ 0.08& 0.71  $\pm$ 0.06& 0.75 $\pm$ 0.05\\ 
      Cook & 0.82 $\pm$ 0.04&  \textbf{0.03} $\pm$ 0.02 & 0.08 $\pm$ 0.08 & 0.38$\pm$ 0.03 & 0.48  $\pm$ 0.16& 0.52 $\pm$ 0.07\\ 
      Laborer & 0.99$ \pm$ 0.01 &  0.09 $\pm$ 0.02& \textbf{0.02}  $\pm$ 0.01 & 0.81 $\pm$ 0.08& 0.81  $\pm$ 0.06& 0.98 $\pm$ 0.03\\ 
      Constr. worker &  1.0 $\pm$ 0.00 & \textbf{0.06} $\pm$ 0.04 & 0.40 $\pm$ 0.06& 0.32 $\pm$ 0.01 & 0.95  $\pm$ 0.01& 1.0 $\pm$ 0.0\\ 
      Developer & 0.90 $\pm$ 0.03&  0.51$\pm$ 0.02 & \textbf{0.14} $\pm$ 0.04 & 0.50 $\pm$ 0.01& 0.74  $\pm$ 0.02& 0.90 $\pm$ 0.04\\ 
      Carpenter & 0.92 $\pm$ 0.05&  0.06 $\pm$ 0.02& \textbf{0.04} $\pm$ 0.09 & 0.52 $\pm$ 0.06& 0.84 $\pm$ 0.01& 0.98 $\pm$ 0.01\\ 
      Manager & 0.54 $\pm$ 0.06& 0.19  $\pm$ 0.07 & \textbf{0.12}$\pm$ 0.03 & 0.38 $\pm$ 0.05& 0.15 $\pm$ 0.01& 0.30 $\pm$ 0.05\\ 
      Lawyer & 0.46 $\pm$ 0.08&  0.30$\pm$ 0.07 & \textbf{0.18} $\pm$ 0.02 & 0.64 $\pm$ 0.03& 0.13 $\pm$ 0.06& 0.52 $\pm$ 0.05\\ 
      Farmer & 0.97 $\pm$ 0.02& 0.41 $\pm$ 0.01 & \textbf{0.18}$\pm$ 0.02 & 0.46 $\pm$ 0.02& 0.58  $\pm$ 0.09& 0.97$\pm$ 0.02\\ 
      Salesperson & 0.60$\pm$ 0.08 & 0.38  $\pm$ 0.05 & 0.14 $\pm$ 0.02& 0.52 $\pm$ 0.05 & 0.18  $\pm$ 0.05& \textbf{0.07}$\pm$ 0.05\\ 
      Physician & 0.62 $\pm$ 0.14& 0.42  $\pm$ 0.01 & \textbf{0.10} $\pm$ 0.07& 0.56 $\pm$ 0.06& 0.36  $\pm$ 0.10& 0.70$\pm$ 0.07\\ 
      Guard &  0.86 $\pm$ 0.02& 0.24  $\pm$ 0.07& \textbf{0.08} $\pm$ 0.04 & 0.30 $\pm$ 0.10& 0.43  $\pm$ 0.12& 0.48$\pm$ 0.06\\ 
      Analyst &  0.58 $\pm$ 0.12&  \textbf{0.20}  $\pm$ 0.07& 0.27 $\pm$ 0.06 & 0.52 $\pm$ 0.03 & 0.24  $\pm$ 0.18& 0.71$\pm$ 0.02\\ 
      Mechanic & 0.99 $\pm$ 0.01&  \textbf{0.23}  $\pm$ 0.08&  0.24 $\pm$ 0.05 & 0.38 $\pm$ 0.09& 0.65  $\pm$ 0.04& 0.92$\pm$ 0.01\\ 
      Sheriff & 0.99 $\pm$ 0.01&  0.10  $\pm$ 0.03 & \textbf{0.04} $\pm$ 0.01& 0.22 $\pm$ 0.05& 0.38 $\pm$ 0.22& 0.82$\pm$ 0.08\\ 
      CEO & 0.87 $\pm$ 0.03&  0.28  $\pm$ 0.03& \textbf{0.20} $\pm$ 0.11 & 0.28$\pm$ 0.04 & 0.25 $\pm$ 0.11& 0.37$\pm$ 0.01\\ 
      Doctor & 0.78 $\pm$ 0.04&  0.20  $\pm$ 0.02& \textbf{0.12} $\pm$ 0.04& 0.58 $\pm$ 0.03& 0.40  $\pm$ 0.02& 0.50$\pm$ 0.04 \\ 
      \bottomrule
    \end{tabular}
    \caption{Biasedness score of debiased concepts randomly selected from WinoBias dataset measured by $\psi$. $\psi = 0$ means perfect debiasing.  Our proposed method shows a consistent preservation among non-debiased occupations. }
    \label{table:biasedness}
  \end{minipage}%
\end{table*}

\section{Additional Qualitative Results}
\begin{figure*}[t]
 \centering
    \includegraphics[width=1\linewidth]{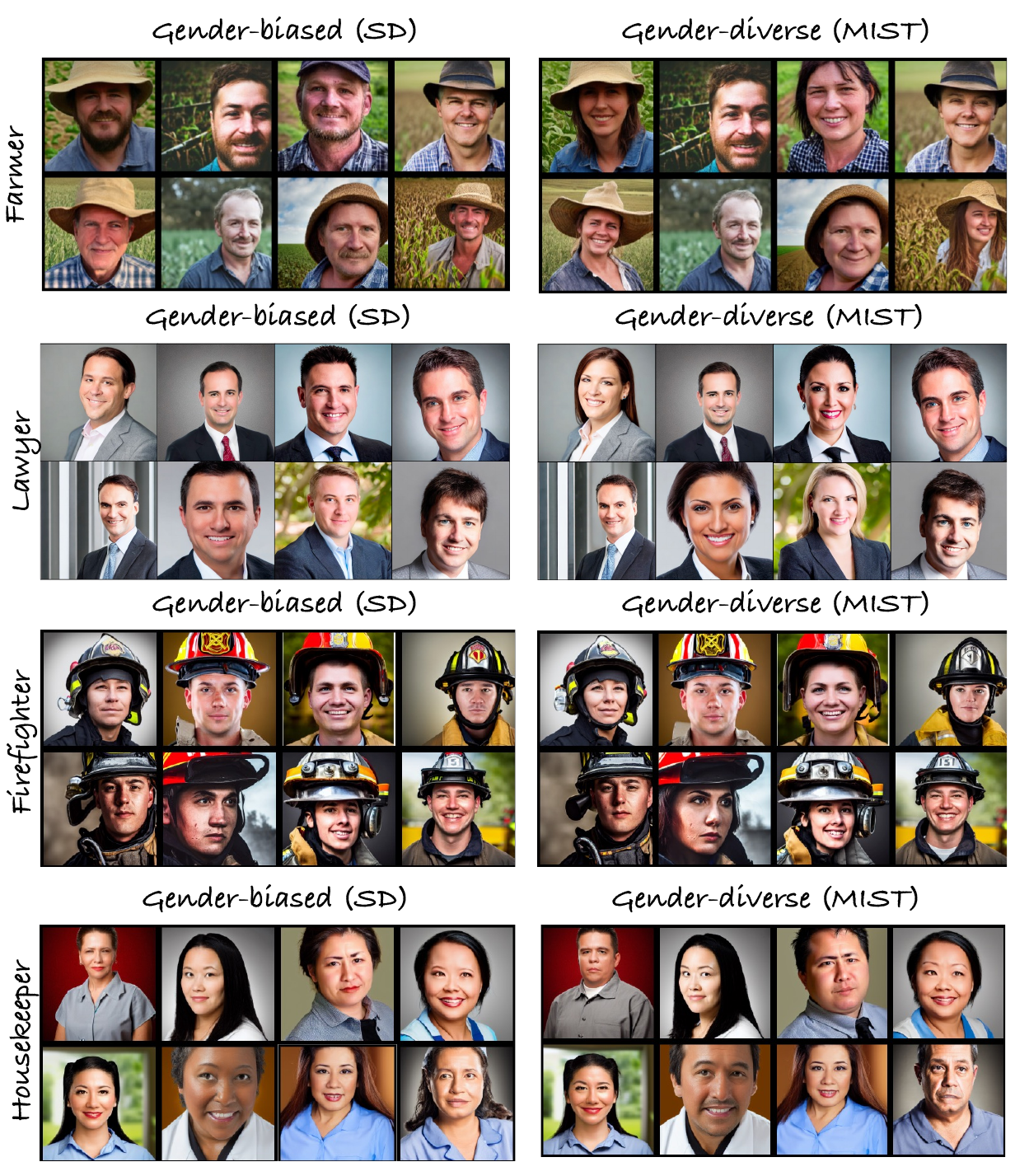}   
 \caption{\textbf{Additional qualitative results for single attribute debiasing with MIST. }}
\hfill
\label{fig:appendix_single}
\end{figure*}

\begin{figure*}[t]
 \centering
    \includegraphics[width=1\linewidth]{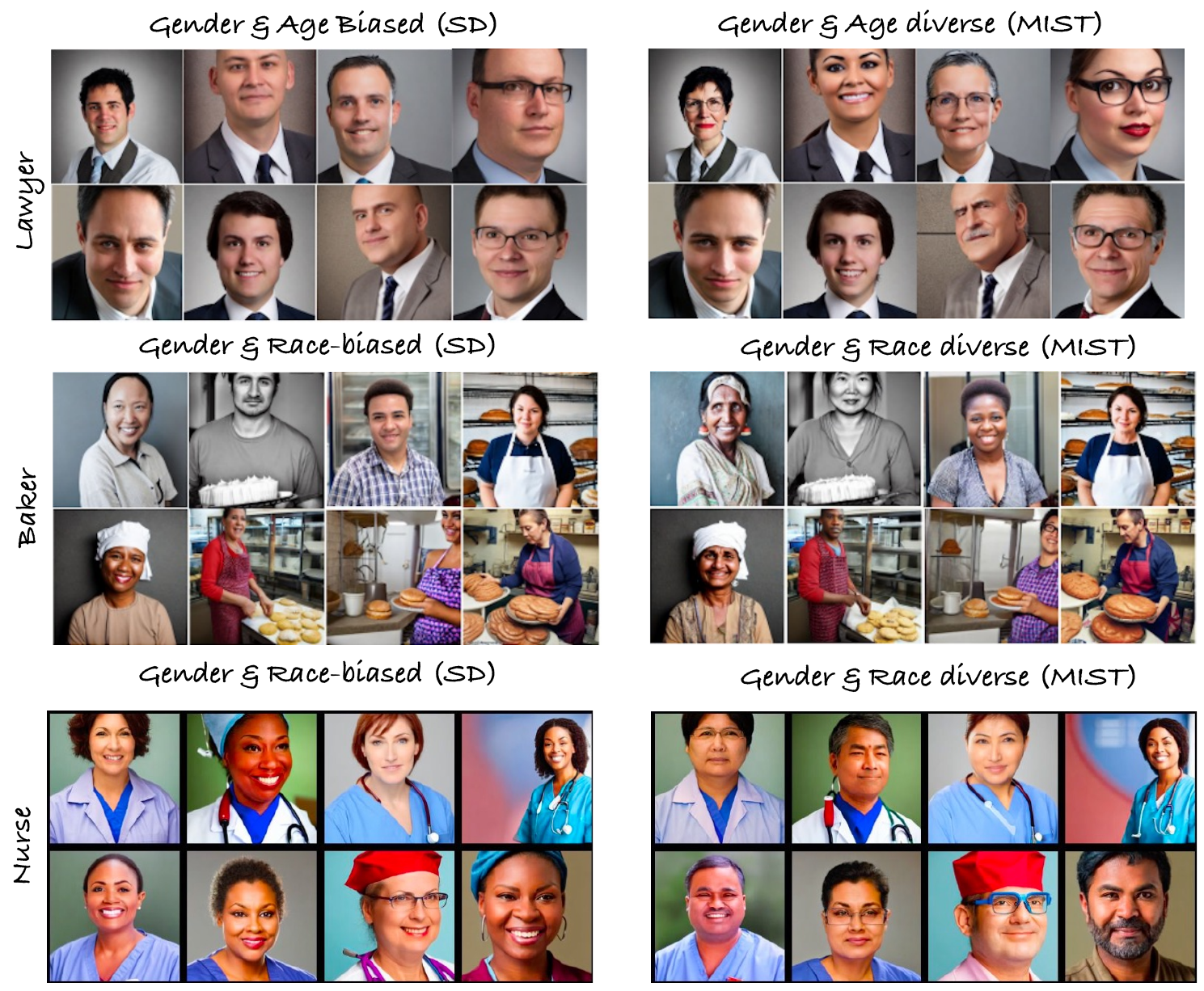}  
 \caption{\textbf{Additional qualitative results for intersectional attribute debiasing with MIST. }}
\hfill
\label{fig:appendix_intersectional}
\end{figure*}

Additional results showcasing both single and combined biases are presented in Fig.  \ref{fig:appendix_single} and Fig. \ref{fig:appendix_intersectional}, respectively.  Our approach effectively addresses single biases, like gender bias, enabling the generation of diverse representations in professions with traditionally low female participation, such as Farmer and Firefighter, as illustrated in Figure \ref{fig:appendix_single}. Conversely, certain professions, such as Housekeeper, which are historically biased towards female representation (shown in Figure \ref{fig:appendix_single}, bottom row), are also diversified by our method, ensuring a more balanced male presence in the debiased versions. Moreover, we provide additional qualitative results for intersectional biases. Our method can successfully handle multiple attributes at the same time, such as Gender \& Race, Gender \& Age, and Gender \& Eyeglasses (see Fig. \ref{fig:appendix_intersectional}). 

